\documentclass{bmvc2k}
\usepackage{amsmath}
\usepackage{amsfonts}
\usepackage{amssymb}
\usepackage{calrsfs}
\usepackage{hyperref}

\DeclareMathAlphabet{\pazocal}{OMS}{zplm}{m}{n}

\title{Global Deconvolutional Networks for Semantic Segmentation}

\addauthor{Vladimir Nekrasov}{nekrasowladimir@unist.ac.kr}{1}
\addauthor{Janghoon Ju}{janghoon.ju@unist.ac.kr}{1}
\addauthor{Jaesik Choi}{jaesik@unist.ac.kr}{1}

\addinstitution{
 Ulsan National Institute of Science and Technology\\
 50 UNIST, Ulju, Ulsan, 44919 Korea
}

\runninghead{Nekrasov \bmvaEtAl}{Global Deconv. Networks for Semantic Segmentation}

\usepackage[english, status=draft]{fixme}
\fxusetheme{color}

\def\etal{\emph{et al}\bmvaOneDot}

\begin{document}

\maketitle

\begin{abstract}
Semantic image segmentation is a principal problem in computer vision, where the aim is to correctly classify each individual pixel of an image into a semantic label. Its widespread use in many areas, including medical imaging and autonomous driving, has fostered extensive research in recent years. Empirical improvements in tackling this task have primarily been motivated by successful exploitation of Convolutional Neural Networks (CNNs) pre-trained for image classification and object recognition. However, the pixel-wise labelling with CNNs has its own unique challenges: (1) an accurate deconvolution, or upsampling, of low-resolution output into a higher-resolution segmentation mask and (2) an inclusion of global information, or context, within locally extracted features. To address these issues, we propose a novel architecture to conduct the equivalent of the deconvolution operation globally and acquire dense predictions. We demonstrate that it leads to improved performance of state-of-the-art semantic segmentation models on the PASCAL VOC 2012 benchmark, reaching $74.0\%$ mean IU accuracy on the test set. 
\end{abstract}

\section{Introduction}
\label{Introduction}
Convolutional Neural Networks \cite{LeCun89} have become an essential part of deep learning models \cite{LeCun2015} designed to tackle a wide range of computer vision tasks including image classification and recognition \cite{KrizhevskySH12, SermanetEZMFL13, SimonyanZ14a, SzegedyLJSRAEVR14, ZeilerF14}, image captioning \cite{KarpathyL15, XuBKCCSZB15, VinyalsTBE15}, object detection \cite{GirshickDDM14, Girshick15, RenHG015}. Recent advances in computing technologies with efficient utilisation of Graphical Processing Units (GPUs), as well as availability of large-scale datasets \cite{DengDSLL009, LinMBHPRDZ14} have been among primary factors in such a rapid rise in CNN popularity.

An adaptation of convolutional network models \cite{LongSD15}, pre-trained for the image classification task, has fostered extensive research on the exploitation of CNNs in semantic image segmentation - a problem of marking (or classifying) each pixel of the image with one of the given semantic labels. Among important applications of this problem are road scene understanding \cite{AlvarezGLL12, BadrinarayananH15, SturgessALT09}, biomedical imaging \cite{RonnebergerFB15, CiresanGGS12}, aerial imaging \cite{KlucknerMRB09, MnihH10}.

Recent breakthrough methods in the area have efficiently and effectively combined neural networks with probabilistic graphical models, such as Conditional Random Fields (CRFs) \cite{ChenPKMY14, LinSRH15, ZhengJRVSDHT15} and Markov Random Fields (MRFs) \cite{LiuLLLT15}. These approaches usually refine per-pixel features extracted by CNNs (so-called `unary potentials') with the help of pairwise similarities between the pixels based on location and colour features, followed by an approximate inference of the obtained fully connected graphical model \cite{KrahenbuhlK11}.

In this work, we address two main challenges of current CNN-based semantic segmentation methods \cite{ChenPKMY14, LongSD15}: an effective deconvolution, or upsampling, of low-resolution output from a neural network; and an inclusion of global information, or context, into existing models without relying on graphical models. Our contribution is twofold: i) we propose a novel approach for performing the deconvolution operation of the encoded signal and ii) demonstrate that this new architecture, called \emph{`Global Deconvolutional Network'}, achieves close to the state-of-the-art performance on semantic segmentation with a simpler architecture and significantly lower number of parameters in comparison to the existing models~\cite{ChenPKMY14,NohHH15,LiuLLLT15}.
\begin{figure*}
\vskip 0.2in
\begin{center}
\centerline{\includegraphics[width=\linewidth]{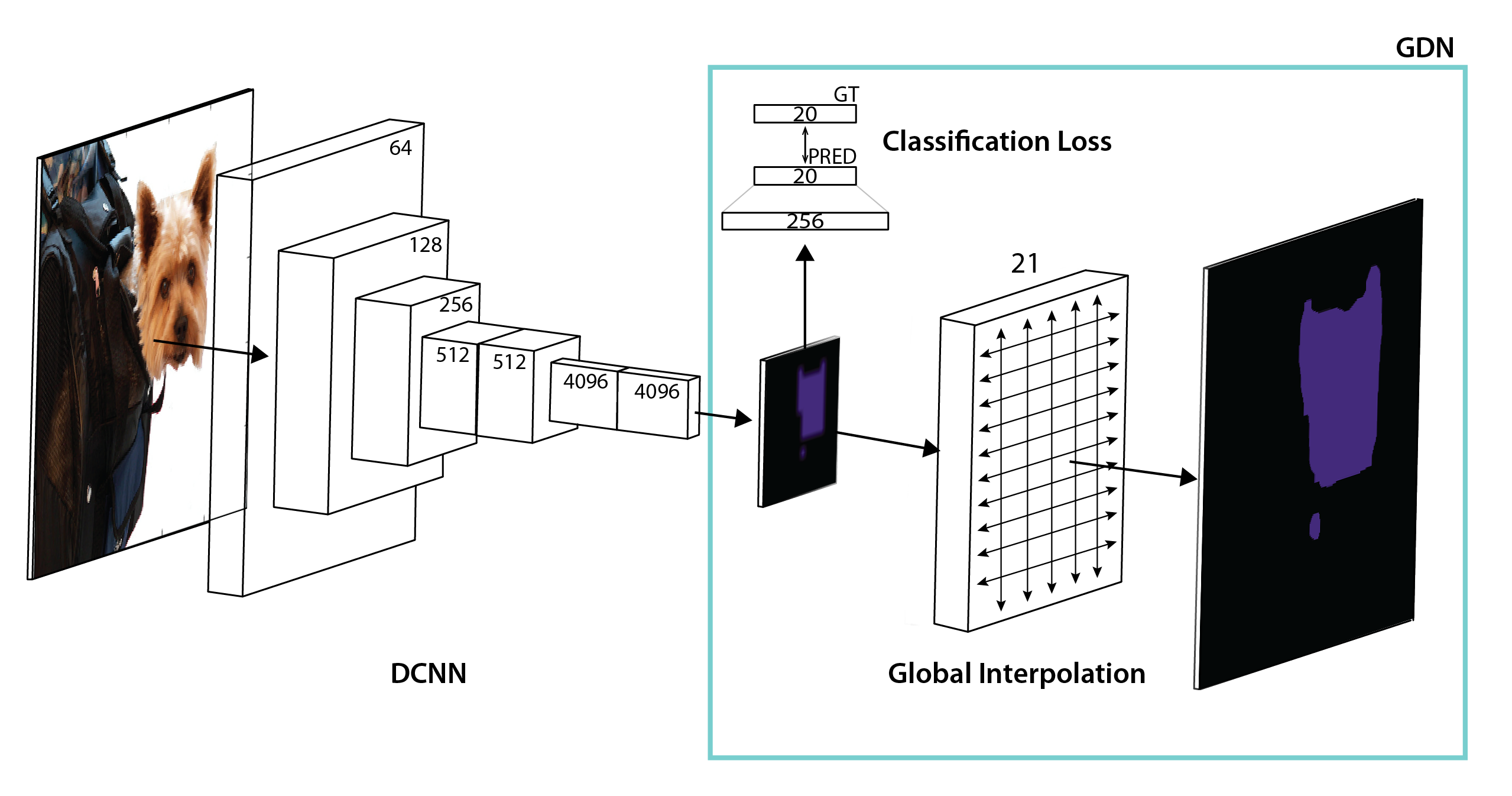}}
\caption{\textbf{Global Deconvolutional Network.} Our adaptation of FCN-32s \cite{LongSD15}. After hierarchical blocks of convolutional-subsampling-nonlinearity layers, we are upsampling the reduced signal with the help of the global interpolation block. In addition to the pixel-wise softmax loss (not shown here), we also use the multi-label classification loss to increase the recognition accuracy.}
\label{architecture}
\end{center}
\vskip -0.2in
\end{figure*} 

The rest of the paper is structured as follows. We briefly explore recent common practices of semantic segmentation models in Section~\ref{Related Work}. Section~\ref{GDN} presents our approach designed to overcome the issues outlined above. Section~\ref{exps} describes the experimental part, including the evaluation results of the proposed method on the popular PASCAL VOC dataset. Finally, Section~\ref{concl} contains conclusions.

\section{Related Work}
\label{Related Work}

Exploitation of fully convolutional neural networks has become ubiquitous in semantic image segmentation ever since the publication of the paper by Long \etal \cite{LongSD15}. Further research has been concerned with the combination of CNNs and probabilistic graphical models \cite{ChenPKMY14, LinSRH15, LiuLLLT15, ZhengJRVSDHT15}, training in the presence of weakly-labelled data \cite{HongNH15, PapandreouCMY15, RussakovskyBFL15}, 
learning an additional (deconvolutional) network \cite{NohHH15}. 

The problem of incorporation of contextual information has been an active research topic in computer vision \cite{RabinovichVGWB07, HeitzK08, DivvalaHHEH09, DoerschGE14, MottaghiCLCLFUY14}.
To some extent, probabilistic graphical models address this issue in semantic segmentation and can be either a) used as a separate post-processing step \cite{ChenPKMY14} or b) trained end-to-end with CNNs \cite{LinSRH15, LiuLLLT15, ZhengJRVSDHT15}.
In setting a) graphical models are unable to refine the parameters of the CNN and thus the errors from the CNN will essentially be presented during post-processing. On the other hand, in b) one need to carefully design the inference part in terms of usual neural networks operations, and it still relies on computing high-dimensional Gaussian kernels \cite{AdamsBD10}. 
Besides that, Yu and Koltun \cite{YuK15} have recently shown that dilated convolution filters are generally applicable and allow to increase the contextual capacity of the network as well.

In terms of improving the deconvolutional part of the network for dense predictions, there has been a prevalence of using information from lower layers: so-called `Skip Architecture' \cite{LongSD15,RonnebergerFB15} and Multi-scale \cite{ChenPKMY14} are two notable examples. 
Noh \etal \cite{NohHH15} proposed to train a separate deconvolutional network to effectively decode information from the original fully convolutional model. While these methods have given better results, all of them contain significantly more parameters than the corresponding baseline models. 

In turn, we propose another approach, called \emph{`Global Deconvolutional Network'}, which includes a global interpolation block with an additional recognition loss, and gives better results than multi-scale and `skip' variants. 
The depiction of our architecture is presented in Figure~\ref{architecture}.

\section{Global Deconvolutional Network}
\label{GDN}
In this section, we describe our approach intended to boost the performance of deep learning models on the semantic segmentation task.
\subsection{Baseline Models}
\label{baseline}
 
As baseline models, we choose two publicly available deep CNN models: FCN-32s\footnote{https://github.com/BVLC/caffe/wiki/Model-Zoo} \cite{LongSD15} and DeepLab\footnote{https://bitbucket.org/deeplab/deeplab-public/} \cite{ChenPKMY14}.
Both of them are based on the VGG 16-layer net \cite{SimonyanZ14a} from the ILSVRC-2014 competition \cite{RussakovskyDSKS15}. This network contains 16 weight layers, including two fully-connected ones, and can be represented as hierarchical stacks of convolutional layers with rectified linear unit non-linearity \cite{GlorotBB11} followed by pooling operations after each stack. The output of the fully-connected layers is fed into a softmax classifier.

For semantic segmentation, where one needs to acquire dense predictions, the fully-connected layers have been replaced by convolution filters followed by a learnable deconvolution or fixed bilinear interpolation to match the original spatial dimensions of the image. The pixel-wise softmax loss represents the objective function.

\subsection{Global Interpolation}
\label{lin_deconv}

The output of multiple blocks of convolutional and pooling layers is an encoded image with severely reduced dimensions. To predict the segmentation mask of the same resolution as the original image, one needs to simultaneously decode and upsample this coarse output. A natural approach is to perform an interpolation. In this work, instead of applying conventional local methods, we devise a learnable global interpolation. 

\begin{figure}[ht]
\vskip 0.2in
\begin{center}
\centerline{\includegraphics[scale=0.4]{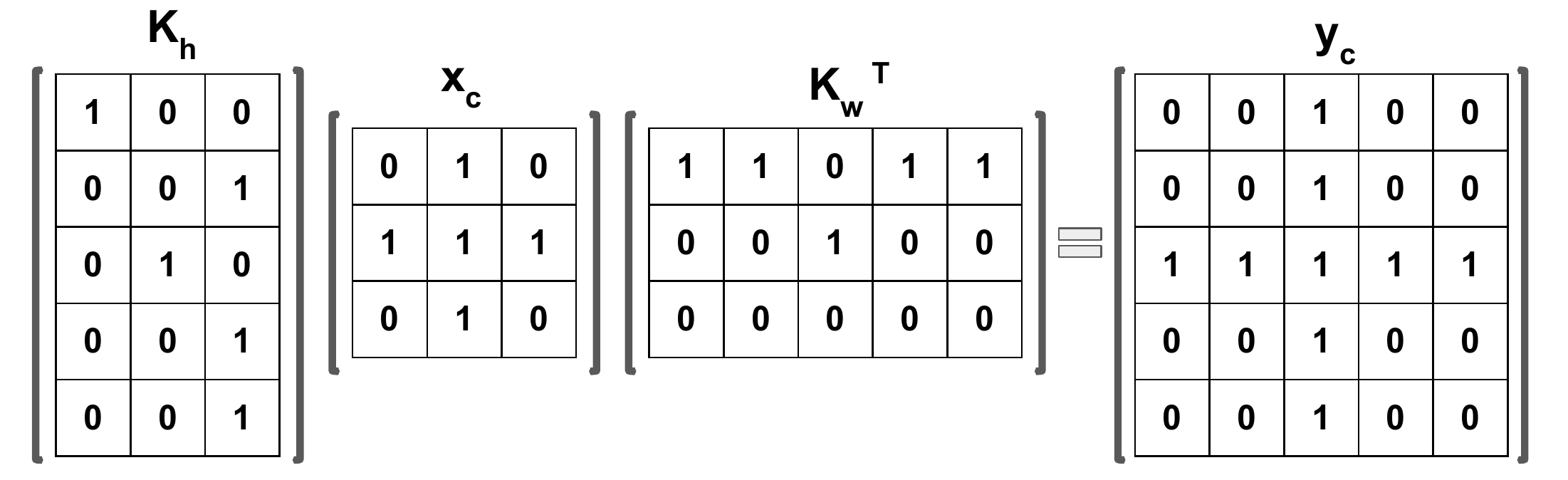}}
\caption{Depiction of Equation~\eqref{eqn1}  for synthetic data.}
\label{equation1}
\end{center}
\vskip -0.2in
\end{figure}

We denote the decoded information of the RGB-image $\mathbf{I}: \mathbf{I}\in{\mathbb{R}^{3\times{H}\times{W}}}$, as $\mathbf{x}:\mathbf{x}\in{\mathbb{R}^{C\times{h}\times{w}}}$, where $C$ represents the number of channels, $h$ and $w$ define the reduced height $H$ and width $W$, respectively. To acquire $\mathbf{y}:\mathbf{y}\in{\mathbb{R}^{C\times{H}\times{W}}}$, an upsampled signal, we apply the following formula:
\begin{equation}
\label{eqn1}
\mathbf{y_{c}}=\mathbf{K_{h}}\mathbf{x_{c}}\mathbf{K_{w}^{\rm T}},
\forall{\mathbf{c}\in{\mathbf{C}}}
\end{equation}
where the matrices $\mathbf{K_{h}}\in{\mathbb{R}^{H\times{h}}}$ and $\mathbf{K_{w}}\in{\mathbb{R}^{W\times{w}}}$ are interpolating each feature map of $\mathbf{x}$ through the corresponding spatial dimensions. Opposite to a simple bilinear interpolation, which operates only on the closest four points, the equation above allows to include much more information on the rectangular grid. An illustrative example can be seen in Figure~\ref{equation1}.

Note that this operation is differentiable, and during the backpropagation algorithm \cite{Rumelhart1986} the derivatives of the pixelwise cross-entropy loss function $\pazocal{L}_{s}$ with respect to the input $\mathbf{x}$ and parameters $\mathbf{K_{h}},\mathbf{K_{w}}$ can be found as follows:
\begin{align}
\label{eqn2}
\begin{gathered}
\frac{\partial \pazocal{L}_{s}}{\partial \mathbf{x_{c}}}=\mathbf{K_{h}^{\rm T}}\frac{\partial \pazocal{L}_{s}}{\partial \mathbf{y_{c}}}\mathbf{K_{w}^{\rm T}}, \\
\frac{\partial \pazocal{L}_{s}}{\partial \mathbf{K_{w}}}=(\frac{\partial \pazocal{L}_{s}}{\partial \mathbf{y_{c}}})^{\mathbf{\rm T}}\mathbf{K_{h}}\mathbf{x_{c}}, \frac{\partial \pazocal{L}_{s}}{\partial \mathbf{K_{h}}}=(\frac{\partial \pazocal{L}_{s}}{\partial \mathbf{y_{c}}})\mathbf{K_{w}}\mathbf{x_{c}^{\rm T}}.
\end{gathered}
\end{align}

We call the operation performed by Equation~\eqref{eqn1} \textit{`global deconvolution'.} We only use this term to underline the fact that we mimic the behaviour of standard deconvolution using a global function; note that our method is not the inverse of the convolution operation and therefore does not represent deconvolution in the strictest sense as, for example, in \cite{ZeilerKTF10}. 

\subsection{Multi-task loss}

It is not uncommon to force intermediate layers of deep learning networks to preserve meaningful and discriminative representations. For example, Szegedy~\etal \cite{SzegedyLJSRAEVR14} appended several auxiliary classifiers to the middle blocks of their model. 
  
As semantic image segmentation essentially comprises image classification as one of its sub-tasks, we append an additional objective function on the top of the coarse output to further improve the model performance on the particular task of classification  (Figure~\ref{architecture}). This supplementary block consists of $3$ fully-connected layers, with the length of the last one being equal to the pre-defined number of possible labels (excluding the background). As there are usually multiple instances of the same label presented in the image, we do not explicitly encode the quantitative components and only denote the presence of a particular class or its absence. The scores from the last layer are transformed with the sigmoid function followed by the multinomial cross entropy loss. 

The loss functions are defined as follows:
\begin{align}
\label{eqn3}
\begin{gathered}
\pazocal{L}_{s}(I, G)=\frac{-1}{|I|}\sum_{i\in{I}}log(\hat{p}_{iG_{i}}), \\
\pazocal{L}_{c}(I, L)=\frac{-1}{|C|}\sum_{c\in{C}}[p_{c}log(\hat{p}_{c})+(1-p_{c})log(1-\hat{p}_{c})],
\end{gathered}
\end{align}
where $\pazocal{L}_{c}$ is the multi-label classification loss; $\pazocal{L}_{s}$ is the pixelwise cross-entropy loss; $I$ is the set of pixels; $G$ is a ground truth map; $C$ is the number of possible labels; $L$ is a ground truth binary vector of length $|C|$; $\hat{p}_{ic}\in[0,1]$ is the softmax probability of pixel $i$ being assigned to class $c$; $p_{c}\in\{0,1\}$ indicates the presence of class $c$ or its absence; $\hat{p}_{c}\in[0,1]$ corresponds to the predicted probability of class $c$ being presented in the image. Note that it is possible to use a weighted sum of the two losses depending on which task's performance we want to optimise.
\\

Overall, each component of the proposed approach aims to capture global information and incorporate it into the network, hence the name \textit{global deconvolutional network}. Besides that, the proposed interpolation also effectively upsamples the coarse output and a nonlinear upsampling can be achieved with the addition of an activation function on the top of the block. The complete architecture of our approach is presented in Figure~\ref{architecture}.

\section{Experiments}
\label{exps}
\subsection{Implementation details} 

We have implemented the proposed methods using Caffe \cite{JiaSDKLGGD14}, the popular deep learning framework. Our training procedure follows the practice of the corresponding baseline models: DeepLab \cite{ChenPKMY14} and FCN \cite{LongSD15}. Both of them employ the VGG-16 net pre-trained on ImageNet \cite{DengDSLL009}. 

We use Stochastic Gradient Descent (SGD) with momentum and train with a minibatch size of 20 images. 
We start the training process with the learning rate equal to $10^{-8}$ and divide it by $10$ when the validation accuracy stops improving. 
We use momentum of $0.9$ and weight decay of $0.0005$.
We initialise all additional layers randomly as in \cite{GlorotB10} and fine-tune them by backpropagation with a lower learning rate before finally training the whole network. 

\subsection{Dataset}

We evaluate performance of the proposed approach on the PASCAL VOC 2012 segmentation benchmark \cite{EveringhamGWWZ10}, which consists of 20 semantic categories and one background category.
Following \cite{HariharanABMM11}, we augment the training data to 8498 images and to 10582 images for the FCN and DeepLab models, respectively. 

The segmentation performance is evaluated by the mean pixel-wise intersection-over-union (mIoU) score \cite{EveringhamGWWZ10}, defined as follows:
\begin{equation}
\label{miou}
mIoU = \frac{1}{|C|}\sum_{c\in{C}}n_{cc}/\left(\sum_{c'\in{C}}{(n_{cc'}+n_{c'c})}-n_{cc}\right),
\end{equation}
where $n_{cc'}$ represents the number of pixels of class $c$ predicted to belong to class $c'$.

First, we conduct all our experiments on the PASCAL VOC \textit{val} set, and then compare the best performing models with their corresponding baseline models on the PASCAL VOC \textit{test} set. As the annotations for the test data are not available, we send our predictions to the PASCAL VOC Evaluation Server.\footnote{http://host.robots.ox.ac.uk/}

\begin{figure*}
\vskip 0.2in
\begin{center}
\centerline{\includegraphics[page=1,height=\dimexpr\textheight-101pt\relax]{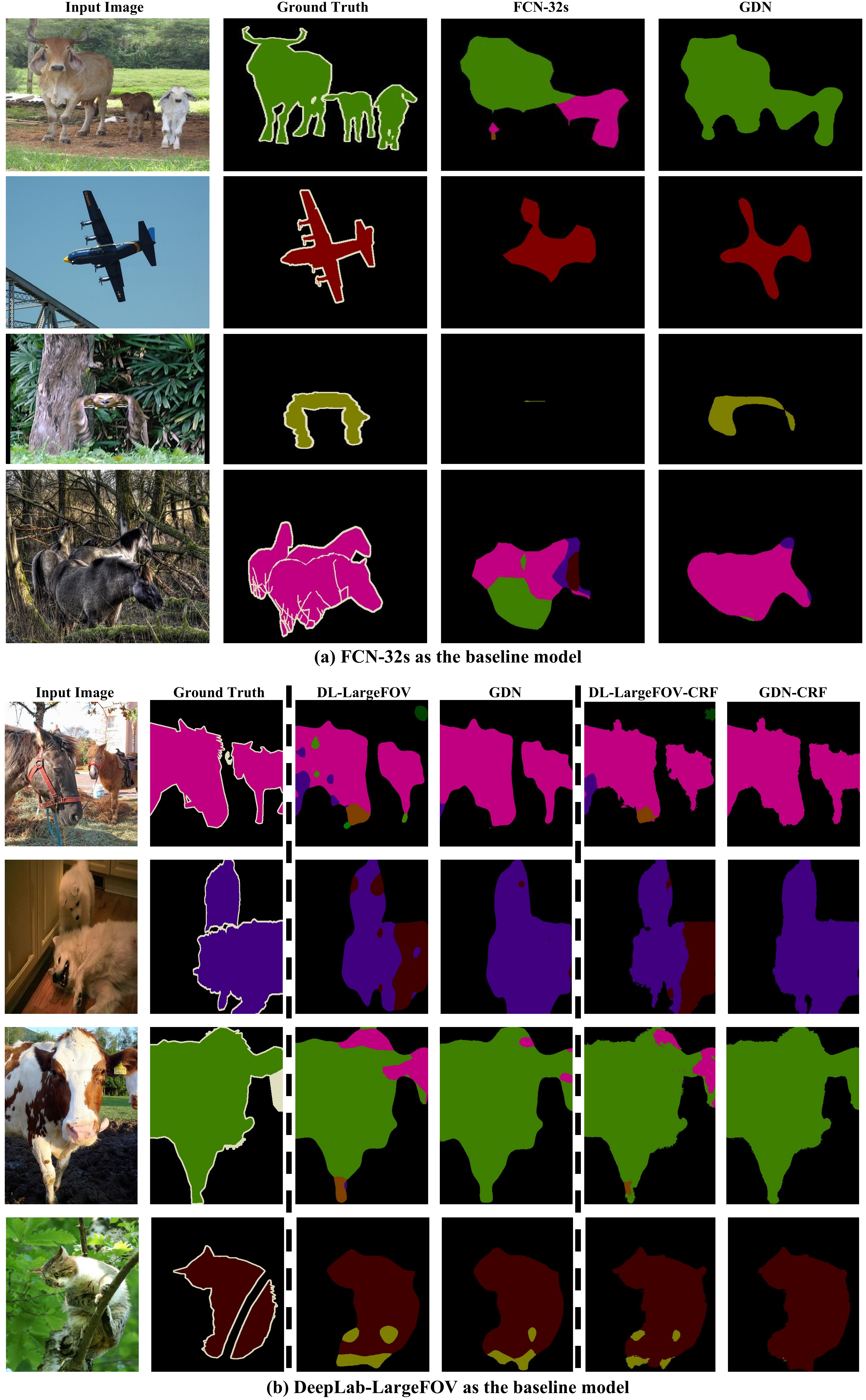}}
\caption{ \textbf{Qualitative results on the validation set}. \textit{(a)} Last column represents our approach, which includes the replacement of standard deconvolution with global deconvolution and the addition of the multi-label classification loss. \textit{(b)} Fourth and sixth columns demonstrate our model, where bilinear interpolation is replaced with global deconvolution. Last two columns also incorporate a conditional random field (CRF) as a post-processing step. \textit{Best viewed in colour.}}
\label{fcn32s}
\end{center}
\vskip -0.2in
\end{figure*}

\subsection{Experiments with FCN-32s}
We conduct several experiments with FCN-32s as a baseline model. During the training stage the images are resized to $500\times 500$.\footnote{This is the maximum value for both the height and width in the PASCAL VOC dataset.} We evaluate all the models on the holdout dataset of 736 images as in \cite{LongSD15}, and send the test results of the best performing ones to the evaluation server.

The original FCN-32s model employs the standard deconvolution operation (also known as \textit{backwards convolution}) to upsample the coarse output. We replace it with the proposed global deconvolution and randomly initialise the new parameters as in \cite{GlorotB10}. We fix the rest of the network to pre-train the added block, and after that train the whole network. 
Global interpolation already improves its baseline model on the validation dataset, as can be seen in Table \ref{table1}.

The baseline model deals with inputs of different sizes via cropping the predicted mask to the same resolution as the corresponding input. Other popular options include either 1) padding or 2) resizing to the fixed input dimensions. In case of global deconvolution, we propose a more elegant solution. Recall that the parameters of this block can be represented as matrices $\mathbf{K_{h}}\in{\mathbb{R}^{H\times{h}}}$ and $\mathbf{K_{w}}\in{\mathbb{R}^{W\times{w}}}$, where ${H}={W}=500$, ${h}={w}=16$ during the training stage. Then, given a test image $\{\mathbf{I}\in{\mathbb{R}^{3\times{\hat{H}}\times{\hat{W}}}}|\hat{W},\hat{H}\leq{500}\}$, we subset the learned matrices to acquire $\mathbf{\hat{K}_{{h}}}\in{\mathbb{R}^{\hat{H}\times{\hat{h}}}}$ and $\mathbf{\hat{K}_{w}}\in{\mathbb{R}^{\hat{W}\times{\hat{w}}}}$ ($\hat{w},\hat{h}\leq{16}$) and proceed with the same operation. To subset, we only leave first $\hat{H},\hat{W}$ rows and $\hat{h},\hat{w}$ columns of the corresponding matrices, and discard all the rest. We have found that this do not affect the final performance of our model.

Next, to increase the recognition accuracy we also append the multi-label classification loss. This slightly improves the validation score in comparison to the baseline model, while the combination with global interpolation gives a further boost in performance (FCN-32s+GDN).   
 
Besides that, we have also conducted additional experiments with FCN-32s, where we insert a fully-connected layer directly after the coarse output (FCN-32s+FC). The idea behind this trick is to allow the network to refine the local predictions based on the information from all the pixels. One drawback in such an approach is the requirement of the fully-connected layers to have the fixed-size input, although the solutions discussed above are also applicable here. Nevertheless, neither of these methods gives satisfactory results during the empirical evaluations. Therefore, we proceed with a slightly different architecture: before appending the fully-connected layer, we first add a spatial pyramid pooling layer \cite{HeZR015}, which produces the output of the same length given an arbitrarily sized input. In particular, we are using a $5$-level pyramid with max-pooling. Though during evaluation this approach alone does not give any improvement in the validation score over the baseline model, its ensemble with the global deconvolution model (FCN-32s+GDN+FC) improves previous results, which indicates that these models may be complementary to each other. 

\begin{table}
\begin{center}
\begin{tabular}{|l || c|}
\hline
Method & mean IoU \\
\hline
FCN-32s \cite{LongSD15}    & 59.4 \\
FCN-32s + Label Loss & 59.8 \\
FCN-32s + Global Interp. & 60.9 \\
FCN-32s + GDN & \textbf{61.2} \\
FCN-32s + GDN + FC & \textbf{62.5} \\  
\hline
DL-LargeFOV \cite{ChenPKMY14}    & 73.3 \\
DL-LargeFOV + Label Loss & 73.9 \\
DL-LargeFOV + Global Interp. & 74.2 \\
DL-LargeFOV + GDN  & \textbf{75.1} \\
\hline
\end{tabular}
\end{center}
\caption{Mean intersection over union accuracy of our approach (GDN), which includes the addition of multi-label classification loss and global interpolation, compared with the baseline model on the reduced validation dataset (for FCN-32s) and on the PASCAL VOC 2012 validation dataset (for DL-LargeFOV).}
\label{table1}
\end{table}

We continue with the evaluation of the best performing models on the test set (Table~\ref{table3}). Both of them improve their baseline model, FCN-32s, and even outperform FCN-8s, another model by Long~\etal~\cite{LongSD15} with the skip-architecture, which combines information from lower layers with the final prediction layer.

Some examples of our approach can be seen in Figure~\ref{fcn32s}.

\subsection{Experiments with DeepLab}

As the next baseline model we consider DeepLab-LargeFOV \cite{ChenPKMY14}. With the help of the \textit{algorithme {\`a}  trous} \cite{holschneider1989real, Shensa92}, the model has a larger Field-of-View (FOV), which results in the finer predictions from the network. Besides that, this model is significantly faster and contains fewer parameters, than the plain modification of the VGG-16 net due to the reduced number of filters of the last two convolutional layers. The model employs simple bilinear interpolation to acquire the output of the same resolution as the input. 

We proceed with the same experiments as for the FCN-32s model, except for the ones involving the fully-connected layer. As DeepLab-LargeFOV has a higher resolution coarse output, the inclusion of the fully-connected layer would result in the weight matrix of several billions parameters. Therefore, we omit these experiments. 

We separately replace the bilinear interpolation with global deconvolution, append the label loss and estimate the joint GDN model. We carry out the same strategy outlined above during the testing stage to deal with variable-size inputs. All the experiments lead to improvements over the baseline model, with GDN showing a significantly higher score on the PASCAL VOC 2012 val set (Table \ref{table1}). 

\begin{figure}[ht]
\vskip 0.2in
\begin{center}
\centerline{\includegraphics[scale=0.25]{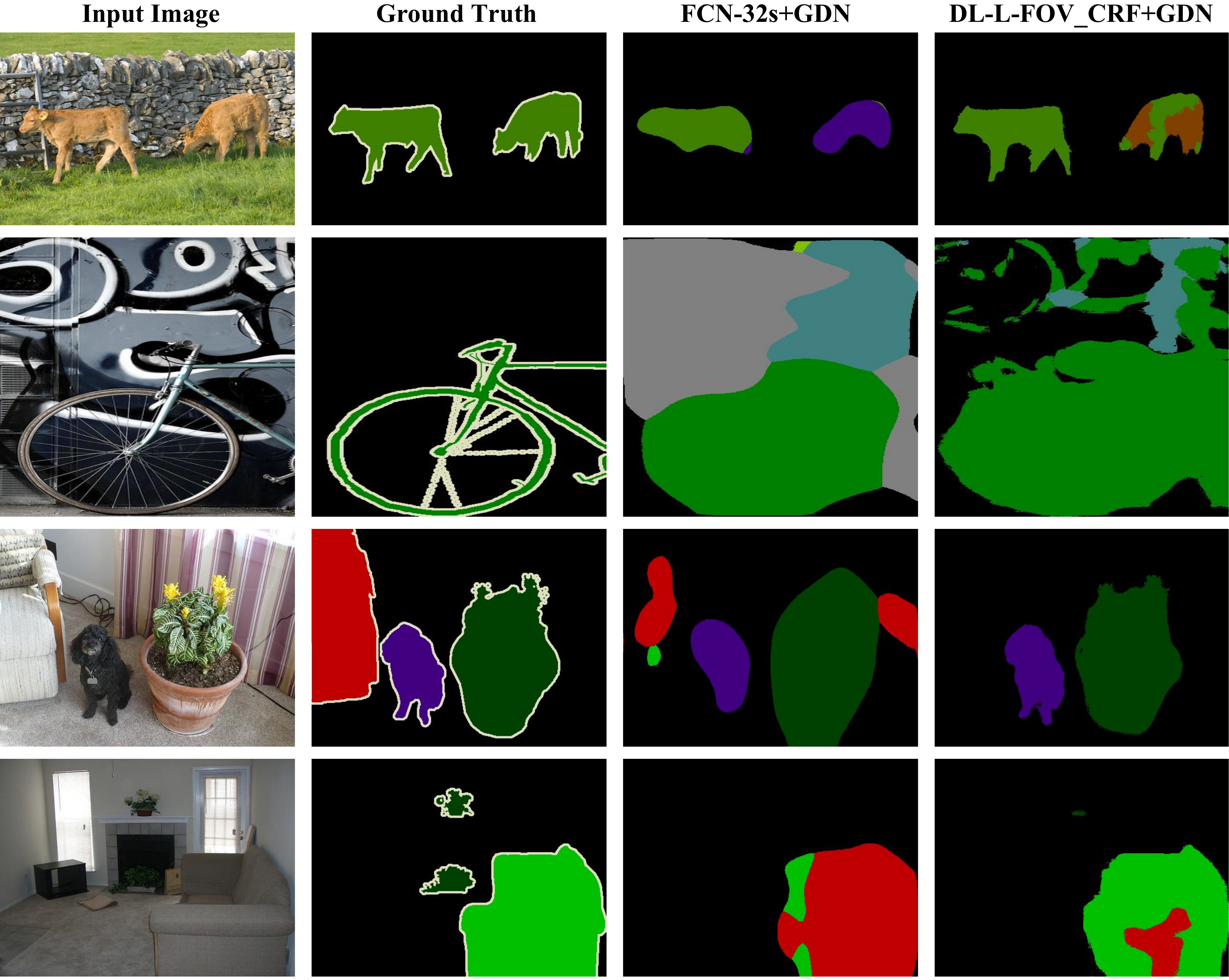}}
\caption{Failure cases of our approach, Global Deconvolutional Network (GDN), on the PASCAL VOC 2012 val set. \textit{Best viewed in colour.} 
}
\label{failure}
\end{center}
\vskip -0.2in
\end{figure}
\addtocounter{footnote}{-1}
\begin{table*}[t]
\setlength{\tabcolsep}{3pt}
\begin{center}
\begin{small}
\begin{sc}
\resizebox{\textwidth}{!}
{\begin{tabular}{|c | *{20}{c} || c |}
\hline
Method & aero & bike & bird & boat & bottle & bus & car & cat & chair & cow & table & dog & horse & mbike & person & plant & sheep & sofa & train & tv & mIoU\\
\hline
FCN-8s~\cite{LongSD15} & \textbf{76.8} & \textbf{34.2} & 68.9 & 49.4 & 60.3 & 75.3 & 74.7 & 77.6 & 21.4 & \textbf{62.5} & 46.8 & 71.8 & \textbf{63.9} & \textbf{76.5} & 73.9 & 45.2 & \textbf{72.4} & 37.4 & 70.9 & 55.1 & 62.20   \\
\textbf{FCN-32s + GDN} & 74.5 & 31.8 & 66.6 & 49.7 & 60.5 & 76.9 & 75.8 & 76.0 & 22.8 & 57.5 & 54.5 & 72.9 & 59.4 & 74.9 & 73.6 & 50.9 & 67.5 & 43.2 & 70.0 & 56.4 & 62.22 \\ 
\textbf{FCN-32s + GDN + FC} & 75.6 & 31.5 & \textbf{69.2} & \textbf{51.6} & \textbf{62.9} & \textbf{78.8} & \textbf{76.7} & \textbf{78.6} & \textbf{24.6} & 61.6 & \textbf{60.3} & \textbf{74.5} & 62.6 & 76.0 & \textbf{74.3} & \textbf{51.4} & 70.6 & \textbf{47.3} & \textbf{73.9} & \textbf{58.3} & \textbf{64.37} \\
\hline
DL-LargeFOV-CRF~\cite{ChenPKMY14} & 83.4 & 36.5 & 82.5 & 62.2 & 66.4 & 85.3 & 78.4 & 83.7 & 30.4 & 72.9 & 60.4 & 78.4 & 75.4 & 82.1 & 79.6 & 58.2 & 81.9 & 48.8 & 73.6 & 63.2 & 70.34 \\
DeconvNet+CRF\_VOC\cite{NohHH15} & 87.8 & 41.9 & 80.6 & 63.9 & 67.3 & 88.1 & 78.4 & 81.3 & 25.9 & 73.7 & 61.2 & 72.0 & 77.0 & 79.9 & 78.7 & 59.5 & 78.3 & 55.0 & 75.2 & 61.5 & 70.50 \\
DL-MSC-LargeFOV-CRF~\cite{ChenPKMY14} & 84.4 & \textbf{54.5} & 81.5 & 63.6 & 65.9 & 85.1 & 79.1 & 83.4 & 30.7 & 74.1 & 59.8 & 79.0 & 76.1 & 83.2 & 80.8 & 59.7 & 82.2 & 50.4 & 73.1 & 63.7 & 71.60 \\
EDeconvNet+CRF\_VOC\cite{NohHH15} & 89.9 & 39.3 & 79.7 & 63.9 & 68.2 & 87.4 & 81.2 & 86.1 & 28.5 & 77.0 & 62.0 & 79.0 & 80.3 & 83.6 & 80.2 & 58.8 & 83.4 & 54.3 & 80.7 & 65.0 & 72.50 \\
\textbf{DL-LargeFOV-CRF + GDN} & 87.9 & 37.8 & \textbf{88.8} & 64.5 & 70.7 & 87.7 & 81.3 & \textbf{87.1} & 32.5 & 76.7 & \textbf{66.7} & 80.4 & 76.6 & 82.2 & 82.3 & 57.9 & 84.5 & 55.9 & 78.5 & 64.2 & 73.21 \\
\textbf{DL-L\_FOV-CRF + GDN\_ENS} & 88.6 & 48.6 & \textbf{88.8} & 64.7 & 70.4 & 87.2 & 81.8 & 86.4 & 32.0 & 77.1 & 64.1 & 80.5 & 78.0 & 84.0 & 83.3 & 59.2 & \textbf{85.9} & 56.8 & 77.9 & 65.0 & \textbf{74.02} \\
Adelaide\_Cont\_CRF\_VOC~\cite{LiuLLLT15} & \textbf{90.6} & 37.6 & 80.0 & \textbf{67.8} & \textbf{74.4} & \textbf{92.0} & \textbf{85.2} & 86.2 & \textbf{39.1} & \textbf{81.2} & 58.9 & \textbf{83.8} & \textbf{83.9} & \textbf{84.3} & \textbf{84.8} & \textbf{62.1} & 83.2 & \textbf{58.2} & \textbf{80.8} & \textbf{72.3} & \textbf{75.30} \\
\hline
\end{tabular}}
\end{sc}
\end{small}
\end{center}
\caption{Mean intersection over union accuracy of our approach (\textbf{GDN}), compared with the competing models on the PASCAL VOC 2012 test set.\protect\footnotemark~Lacking the results of FCN-32s on the test set, we thus compare it directly with a more powerful model, FCN-8s \cite{LongSD15}. All the methods presented here do not use MS COCO dataset.}
\label{table3}
\vskip -0.1in
\end{table*}

The DeepLab-LargeFOV model also incorporates a fully-connected CRF \cite{LaffertyMP01, KohliLT09, KrahenbuhlK11} as a post-processing step. 
To set the parameters of the fully connected CRF, we employ the same method of cross-validation as in \cite{ChenPKMY14} on a subset of the validation data. Then we send our best performing model enriched by CRF to the evaluation server. 
On the PASCAL VOC 2012 test set our single model (DL-LargeFOV-CRF+GDN) achieves $73.2\%$ mIoU, a significant improvement over the baseline model (around $2.9\%$), and even excels the multiscale DeepLab-MSc-LargeFOV model by $1.6\%$ (Table~\ref{table3}); the predictions averaged across our several models (DL-L\_FOV-CRF+GDN\_ENS) give a further improvement of $0.8\%$, showing a competitive score to the models that do not exploit the Microsoft COCO dataset~\cite{LinMBHPRDZ14}.\\
As is the case with the FCN-32s model, we obtain performance on par with the multi-resolution variant using a much simpler architecture. Moreover, our single CRF-equipped global deconvolutional network (DL-LargeFOV-CRF+GDN) even surpasses the results of the competing approach (DeconvNet+CRF~\cite{NohHH15})  by $2.7\%$, where the deconvolutional part of the network contains significantly more parameters: almost 126M compared to less than 70K of global deconvolution; in case of ensembles, the improvement is over $1.5\%$.

The illustrative examples are presented in Figures~\ref{fcn32s} and~\ref{failure}.



\section{Conclusion} 
\label{concl}
In this paper we addressed two important problems of semantic image segmentation: an upsampling of the low-resolution output from the network and refinement of this coarse output, incorporating global information and the additional classification loss. We proposed a novel approach, \textit{global deconvolution}, to acquire the output of the same size as the input for images of variable resolutions. We showed that \textit{global deconvolution} effectively replaces standard approaches, and can easily be trained in a straightforward manner. 

On the benchmark competition, PASCAL VOC 2012, we showed that the proposed approach outperforms the results of the baseline models. Furthermore, our method even surpasses the performance of more powerful multi-resolution models, which combine information from several blocks of the deep neural network. \\

\noindent
\textbf{Acknowledgements}
The authors would like to thank the anonymous reviewers for their helpful and constructive comments, and Gaee Kim for making Fig.~\ref{architecture}.
This work is supported by the Ministry of Science, ICT \& Future Planning (MSIP), Korea, under Basic
Science Research Program through the National Research Foundation of Korea (NRF) grant (NRF-2014R1A1A1002662), under the ITRC (Information Technology Research Center) support program (IITP-2016-R2720-16-0007) supervised by the IITP (Institute for Information \& communications Technology Promotion) and under NIPA (National IT Industry Promotion Agency) program (NIPA-S0415-15-1004).

\footnotetext{\url{http://host.robots.ox.ac.uk:8080/leaderboard/displaylb.php?challengeid=11&compid=6}}%

\bibliography{egbib}
\end{document}